\documentclass[pdflatex,sn-mathphys-num]{sn-jnl}

\usepackage{graphicx}%
\usepackage{CJKutf8}
\usepackage{multirow}%
\usepackage{amsmath,amssymb,amsfonts}%
\usepackage{amsthm}%
\usepackage{mathrsfs}%
\usepackage[title]{appendix}%
\usepackage{xcolor}%
\usepackage{textcomp}%
\usepackage{manyfoot}%
\usepackage{booktabs}%
\usepackage{algorithm}%
\usepackage{algorithmicx}%
\usepackage{algpseudocode}%
\usepackage{listings}%

\begin{document}
\begin{CJK}{UTF8}{gbsn}

\title[Article Title]{\textbf{The high dimensional psychological profile and cultural bias of ChatGPT}
}


\author[1]{\fnm{Hang} \sur{Yuan}}\equalcont{These authors contributed equally to this work.}

\author[1]{\fnm{Zhongyue} \sur{Che}}\equalcont{These authors contributed equally to this work.}

\author[1]{\fnm{Shao} \sur{Li}}\equalcont{These authors contributed equally to this work.}

\author[1]{\fnm{Yue} \sur{Zhang}}\equalcont{These authors contributed equally to this work.}

\author[2]{\fnm{Xiaomeng} \sur{Hu}}

\author*[1]{\fnm{Siyang} \sur{Luo}}\email{ljc520ida@163.com/Luosy6@mail.sysu.edu.cn}

\affil[1]{\orgdiv{Department of Psychology}, \orgname{Guangdong Provincial Key Laboratory of Social Cognitive Neuroscience and Mental Health},\orgname{Guangdong Provincial Key Laboratory of Brain Function and Disease}, \orgaddress{\orgname{Sun Yat-Sen University}, \city{Guangzhou} \postcode{510006}, \country{China}}}

\affil[2]{\orgdiv{Department of Psychology}, \orgname{Renmin University of China}, \orgaddress{\city{Beijing} \postcode{100871},  \country{China}}}


\abstract{Given the rapid advancement of large-scale language models, artificial intelligence (AI) models, like ChatGPT, are playing an increasingly prominent role in human society. However, to ensure that artificial intelligence models benefit human society, we must first fully understand the similarities and differences between the human-like characteristics exhibited by artificial intelligence models and real humans, as well as the cultural stereotypes and biases that artificial intelligence models may exhibit in the process of interacting with humans. This study first measured ChatGPT in 84 dimensions of psychological characteristics, revealing differences between ChatGPT and human norms in most dimensions as well as in high-dimensional psychological representations. Additionally, through the measurement of ChatGPT in 13 dimensions of cultural values, it was revealed that ChatGPT's cultural value patterns are dissimilar to those of various countries/regions worldwide. Finally, an analysis of ChatGPT's performance in eight decision-making tasks involving interactions with humans from different countries/regions revealed that ChatGPT exhibits clear cultural stereotypes in most decision-making tasks and shows significant cultural bias in third-party punishment and ultimatum games. The findings indicate that, compared to humans, ChatGPT exhibits a distinct psychological profile and cultural value orientation, and it also shows cultural biases and stereotypes in interpersonal decision-making. Future research endeavors should emphasize enhanced technical oversight and augmented transparency in the database and algorithmic training procedures to foster more efficient cross-cultural communication and mitigate social disparities.}




\maketitle
\newpage
\section{Introduction}\label{sec1}

\quad Modern artificial intelligence has been applied to realize various human-computer interaction functions, including dialogues, providing suggestions, composing poems, and proving theorems. This highly humanized performance has garnered widespread public attention \cite{r1}, \cite{shackelford2023we}. Previous studies have primarily examined the degree of humanization of ChatGPT from algorithmic and data-centric perspectives, investigating whether ChatGPT can understand and process information akin to humans\cite{binz2023using}, \cite{mei2024turing}, as well as the potential advantages and challenges associated with such humanization\cite{acerbi2023large}, \cite{bubeck2023sparks}. However, these studies have often focused on singular dimensions, primarily assessing ChatGPT's performance while overlooking its comprehensive psychological and cognitive attributes. Therefore, gaining a deeper understanding of its overall human-like mental representation for efficient human-computer interaction is imperative. Furthermore, considering the cultural values ingrained in the training data, ChatGPT may have inadvertently acquired certain cultural biases or stereotypes during its training. Hence, this study utilizes multidimensional indicators to systematically analyze the cultural values and potential biases and stereotypes inherent in ChatGPT.

Measuring ChatGPT's psychological profile aids in understanding its patterns of interaction with humans. While existing research has primarily focused on ChatGPT's personality characteristics, laying a foundation for understanding this complex system, it has often overlooked its multidimensional psychological essence. Psychological characteristics encompass a broad range of aspects, including personality, emotion, cognition, motivation, and values, all of which interact and collectively shape an individual's psychological profile. However, current research tends to be confined to the analysis of personality traits, such as the Dark Triad or the Big Five personality model, neglecting other equally significant dimensions\cite{mei2024turing},\cite{chen2023emergence},\cite{rutinowski2023self}. As a highly intricate artificial intelligence technology, ChatGPT's psychological characteristics extend beyond personality alone; its complexity and adaptability in language processing, context understanding, and response generation are also crucial for assessing its psychological makeup—for instance, ChatGPT's capacity to demonstrate empathy, humor, and emotional expression during conversations. Therefore, to comprehensively uncover the "psychological profile" of ChatGPT, this study adopts multidimensional measurement indicators to evaluate its performance across key dimensions such as personality characteristics, emotional states, autonomy and subjective life and career perception. This comprehensive assessment reveals ChatGPT's internal psychological and cognitive traits and constructs a holistic mental model of ChatGPT across various dimensions. Through this approach, we gain a more accurate understanding of ChatGPT's complexity, assess its potential to model human psychology and cognition, and provide deeper insights into the future development of artificial intelligence.

When exploring the comprehensive psychological profile of ChatGPT, research into its underlying cultural factors is crucial. Studies have shown that individuals from different cultural backgrounds exhibit unique psychological characteristics and behavioral patterns from the values and societal norms emphasized within their respective cultures\cite{mccrae2005universal},\cite{henrich2001cooperation},\cite{oheigeartaigh2020overcoming}. ÓhÉigeartaigh et al., (2020) have pointed out that AI systems often reflect the dominant cultural values in their training data when making moral and ethical decisions\cite{oheigeartaigh2020overcoming}. Therefore, ChatGPT likely possesses inherent cultural values. If ChatGPT exhibits cultural values different from those of the users, it may elicit resistance or distrust from users\cite{bender2021dangers}.

In addition, the cultural values embodied by ChatGPT can contribute to the formation of generalized cognition among different cultural groups\cite{ornelas2023redefining}. This generalized cognition often leads to the formation of biases and stereotypes, where ChatGPT may develop overly simplistic views of specific cultural groups based on limited sources of information. Therefore, societal cultural biases typically lack explicit oppositional norms and are more challenging to identify than other forms, such as gender or racial biases\cite{nosek2002harvesting},\cite{kleinman2016anthropology}. These biases challenge achieving equality in various fields, including healthcare and education. ChatGPT has been widely applied across different domains and provides recommendations for individuals working in various fields. However, if ChatGPT's responses exhibit cultural biases, it not only compromises the neutrality and fairness of interactions but also exacerbates misunderstandings, social divisions, and even group conflicts\cite{olli2023cultural},\cite{yang2023group},\cite{pursiainen2022cultural}. Previous research on cultural biases often relies on questionnaire results, but participants may exhibit self-report biases (such as embellishing or concealing their biases and stereotypes) when completing these questionnaires, thereby affecting the authenticity of the results. Therefore, this study will employ classic behavioral decision paradigms, requiring ChatGPT to simulate individuals from different countries' participation in decision-making to explore its biases towards different national cultures more intuitively.

Furthermore, in addition to looking at ChatGPT's performance or characteristics in each dimension, it may be more important to understand ChatGPT's human-like high-dimensional representation model in general. As shown in Figure 1, even if ChatGPT shows similarity to humans in every dimension, it may still be significantly different from real humans in the overall pattern. We explore these high dimensional patterns of ChatGPT from the perspective of representational similarity. Representational similarity analysis (RSA) initially originated in systems neuroscience and is a specialized multivariate pattern analysis method that reflects the representation of concepts or stimulus units in a high-dimensional space through pairwise comparisons\cite{haxby2014decoding}. Previous studies have often explored the humanness of ChatGPT on a single dimension, such as the Big Five personality traits, neglecting the multidimensionality and complexity of psychological characteristics. Studying psychological or behavioral variables solely from a single dimension is insufficient to capture the full extent of humanness exhibited by ChatGPT, thus necessitating the integration of psychological traits across different dimensions. One advantage of RSA is its ability to extract the degree of association between different dimensions, thereby constructing global feature patterns of object variable\cite{yuanRSA}. Therefore, RSA analysis can help us integrate ChatGPT's overall psychological profile. Additionally, RSA enables the quantification of global pattern information and objective comparisons of pattern features across different scales. Consequently, we can directly compare the differences between ChatGPT and data from different scales regarding overall patterns. In summary, employing pattern analysis may offer us novel and complementary perspectives.

This study aims to establish a comprehensive psychological profile of ChatGPT by examining three aspects: psychological traits, cultural values, and cultural biases and stereotypes. Firstly, multiple psychological trait questionnaires are utilized to measure the multidimensional psychological characteristics of ChatGPT and compare them with human subjects. Subsequently, the values of ChatGPT across different cultural dimensions are measured to explore its value preferences. Additionally, through complex decision-making experiments, the decision tendencies of ChatGPT in different cultural contexts are observed, analyzing its underlying cultural biases and stereotypes. Furthermore, based on RSA analysis, the study explores ChatGPT's psychological profile and overall value patterns.

\section{The psychological profile of ChatGPT}\label{sec2}

\quad To gain a comprehensive understanding of the personality traits exhibited by ChatGPT, a diverse range of psychological scales was employed to assess its personality across multiple dimensions. A systematic search was conducted within the Web of Science database, explicitly focusing on peer-reviewed literature. The search strategy included specific terms related to perception, ability, personality traits, self, interpersonal interactions, life meaning, social well-being, distress, work, affective/emotional state, cognition, and motivation. This meticulous and thorough search identified 44 questionnaires that met the predetermined inclusion and exclusion criteria.

The identified questionnaires were subsequently categorized into four distinct domains: individual characteristics, social interaction, career management, and life experience (see supplementary information for details). The questionnaires were administered to ChatGPT in a standardized format to ensure methodological rigor. Scores were calculated for each scale based on the responses provided by ChatGPT. And then the comparisons between ChatGPT's scores and the human norm were conducted. Figure 2 displays the norm scores for the scale, the confidence intervals for the norm, and the scores obtained from ChatGPT.

Regarding individual characteristics, in terms of autonomy, ChatGPT exhibited higher scores than humans in self-control, indicating superior behavioral regulation (\textit{t}(1138) = -70.59, \textit{p} \(<\) 0.001). However, it obtained lower scores than humans in self-perception and control over the external environment, such as perceived personal control, perceived interpersonal control and perceived social-political control (\textit{t}s \(>\) 3.37,\textit{ p}s \(<\) 0.001), suggesting a perception of reduced influence and control over the behavior and social systems of others. In terms of moral judgment, ChatGPT scored lower than humans in sanctity (\textit{t}(1200) = 25.91, \textit{p} \(<\) 0.001) but higher in loyalty, fairness, care, and authority (\textit{t}s \(<\) -18.97, \textit{p}s \(<\) 0.001), indicating a relatively diminished emphasis on religious beliefs. Concerning metacognition, ChatGPT obtained lower scores than humans in meta mind-controlling needs, meta positive beliefs, and authenticity (\textit{t}s \(>\) 2.71, \textit{p}s \(<\) 0.01), while scoring higher in meta cognitive confidence, meta uncontrollability, and meta cognitive self-consciousness (\textit{t}s \(<\) -11.17,\textit{ p}s \(<\) 0.001). This suggests that ChatGPT demonstrates heightened confidence in its cognitive abilities but shows limited control over its thought processes and a reluctance to reveal its true self. In terms of personality traits, ChatGPT scored higher than humans in grit, narcissism, and self-enhancing humor style (\textit{t}s \(<\) -7.92, \textit{p} \(<\) 0.001) while scoring significantly lower in trait hope, psychopathy, self-defeating humor style, aggressive humor style, and Machiavellianism (\textit{t}s \(>\) 12.40, \textit{p}s \(<\) 0.001). Additionally, ChatGPT did not significantly differ from humans in affiliative humor style (\textit{t}(825) = -1.24, \textit{p} = 0.217). On the emotional, ChatGPT scored lower than humans in social anxiety, self-compassion, loneliness, and gratitude (\textit{ts} \(>\) 7.08, \textit{ps} \(<\) 0.001). However, its scores in emotional empathy, cognitive empathy, and self-stigma of seeking help were significantly higher than those of humans, reflecting a relatively optimistic, cheerful, and approachable character (\textit{t}s \(<\) -27.34, \textit{p}s \(<\) 0.001).

Concerning social interaction, within the dimensions related to interpersonal communication, ChatGPT showed significant differences from humans, except in interpersonal avoidance motivation (\textit{t}(584) = 0.17, \textit{p} = 0.868). However, it obtained lower scores than humans in interpersonal trust (\textit{t}(672) = 4.40, \textit{p} \(<\) 0.001) while scoring higher than humans in interpersonal anxiety, empathic concern for others, perspective taking, identity tendency of fictional characters, interpersonal revenge motivation, and sociability (\textit{t}s \(<\) -16.22, \textit{p}s \(<\) 0.001). In terms of social adaptation, ChatGPT achieved higher scores than humans in social desirability, perceived social support, perceived discrimination, mental resilience, social comparison, culture adaptation, and social referencing (\textit{t}s \(<\) -3.75, \textit{p}s \(<\) 0.001), thus exhibiting heightened adaptability, active integration, and a willingness to conform to social expectations.

Within the occupational domain, with respect to working experience, ChatGPT obtained lower scores than humans in need for autonomy, need for relatedness, and need for competence (\textit{ts} \(>\) 50.16, \textit{ps} \(<\) 0.001). Conversely, it scored higher than humans in dimensions associated with job burnout, including personal burnout, work-related burnout, and client-related burnout (\textit{ts} \(<\) -39.42,\textit{ ps} \(<\) 0.001), indicating a potential inclination towards work fatigue. However, the score on the experience was higher (\textit{t}(4321) = -437.33, \textit{p} \(<\) 0.001), suggesting more positive work experiences. Notably, there was no significant difference between ChatGPT and humans regarding work focus (\textit{t}(198) = -0.21, \textit{p} = 0.831).

About job motivation, ChatGPT achieved lower scores in work intrinsic motivation and lack of work motivation (\textit{t}s \(>\) 7.64; \textit{p}s \(<\) 0.001) while scoring higher in dimensions related to external work motivation, such as external regulation of work, introjected regulation of work, integrated regulation of work, and identified regulation of work (\textit{t}s \(<\) -18.23, \textit{p}s \(<\) 0.001). This indicates that ChatGPT's work motivation relies more heavily on external factors.

In terms of occupational adaptation, ChatGPT scored low in decreasing hindering job demands (\textit{t}(2054) = 7.77, \textit{p} \(<\) 0.001) and higher in dimensions associated with increasing structural job resources, increasing social job resources, and increasing challenging job demands (\textit{t}s \(<\) -37.78, \textit{p}s \(<\) 0.001). This suggests that ChatGPT is more inclined than humans to improve the work environment and increase job satisfaction actively.

As for life experience, in terms of attitude towards life, ChatGPT obtained higher scores in the presence of life meaning and perfectionism (\textit{t}s \(<\) -49.73, \textit{p}s \(<\) 0.001) but lower scores in conspiracy mentality, compassion, growth mindset, searching for life meaning, reflective and cognition of life meaning (\textit{t}s \(>\) 19.90, \textit{p}s \(<\) 0.001). This suggests that ChatGPT may have limitations in accurately perceiving life events. Despite this, ChatGPT tends to follow the rules strictly, display more significant perfectionism, and exhibit lower levels of conspiracy thinking due to its heightened self-control and adherence to social expectations. However, compared to humans, ChatGPT exhibits a lower tendency towards a growth mindset, suggesting a belief in fixed and unchangeable traits and a reduced willingness to embrace novelty.

About life satisfaction, ChatGPT obtained higher scores in well-being of positive relations, happiness, and pleasure but also scored higher in psychological distress (\textit{t}s \(<\) -31.83, \textit{p}s \(<\) 0.001). It demonstrates both elevated levels of happiness and psychological distress. ChatGPT scored higher than humans in satisfaction with positive relationships (\textit{t}(1191) = -9.38, \textit{p}s \(<\) 0.001) while displaying significantly lower satisfaction in the well-being of environmental mastery, well-being of self-acceptance, well-being of personal growth, and well-being of autonomy (\textit{t}s \(>\) 13.90, \textit{p}s \(<\) 0.001). This discrepancy may arise from ChatGPT finding satisfaction primarily in interpersonal connections while exhibiting lower satisfaction in other aspects compared to the human norm, leading to heightened psychological distress. There was no significant difference between ChatGPT and humans in well-being of purpose in life (\textit{t}(1191) = -1.18, \textit{p} = 0.239) (See supplementary Table S1 for details).

As illustrated in Figure 3, the analysis of representational similarity revealed that, at a broad level, ChatGPT exhibited considerable similarity with human beings in various dimensions of psychological characteristics (\textit{r} = 0.12, \textit{p} = 0.009). Regarding sub-dimensions, ChatGPT, and human psychological characteristics displayed significantly similar representation patterns in social interaction (\textit{r} = 0.57, \textit{p} = 0.033). However, regarding personal characteristics, career management, and life experience, ChatGPT diverged from human psychological traits (\textit{r }= 0.004, \textit{p} = 0.479; \textit{r} = 0.20, \textit{p} = 0.062; \textit{r} = 0.01, \textit{p} = 0.369). This indicates that ChatGPT and human beings adopt distinct representation modes regarding personal characteristics, career management, and life experience.

ChatGPT can be characterized as an entity with strong social skills, robust self-control, proactive work tendencies, and an optimistic worldview. However, it exhibits a weaker sense of self and environment. Although it acknowledges the existence of meaning, it does not actively seek to explore its significance. It tends to conform to certain social norms. The exhibited personality traits bear some resemblance to human patterns, with a more pronounced similarity observed in traits related to social interaction.

\section{The culture value orientation of ChatGPT}\label{sec3}

\quad To explore ChatGPT’s cultural bias, we selected 13 common cultural values and measured them using scales. At the same time, we collected data on cultural values in multiple countries/regions to compare with ChatGPT scores.(See Table S2 for details) Figure 4 shows the scores of ChatGPT and different countries/regions in each dimension of cultural values. In terms of independent-interdependent self-construction, ChatGPT ranks at 100\% for independent self-construction and 36.67\% for interdependent self-construction, indicating a stronger tendency towards interdependent self-construction. This implies that ChatGPT tends to construct self-awareness from a social relational perspective. In the gender differentiation/egalitarianism dimension, ChatGPT scores first among 60 regions of 58 countries, suggesting a higher inclination towards gender equality than the overall level. 

Furthermore, in the dimension of assertiveness, ChatGPT ranks at 100\%, emphasizing interpersonal relationships and cooperation. The score ranking of gender differentiation/egalitarianism and assertiveness of ChatGPT shows that ChatGPT is more feminine. Additionally, in the power distance value, ChatGPT’s score ranks at 1.61\%, and in relationship mobility, ChatGPT’s score ranks at 10\%, indicating that ChatGPT values equality and freedom in interpersonal communication. In each cultural dimension, the measurement method for the cultural values of ChatGPT is consistent with Study 1.

In the dimension of individualism-collectivism cultural values, ChatGPT's collectivism level ranks at 16.92\%, and in the GLOBE’s institutional and in-group collectivism cultural values scores rank at 1.61\% and 17.74\%, respectively, which is high-level among all countries/regions. These results suggest that in Hofstede’s and GLOBE’s dimensions of individualism-collectivism national culture, ChatGPT exhibits a higher tendency towards collectivism.

ChatGPT ranked first in both humane orientation and performance orientation, indicating that it not only pays attention to treating others well but also encourages others to achieve achievements. In addition, ChatGPT ranks 96.24\% out of 60 countries/regions in the uncertainty avoidance value, indicating a strong tendency towards risk avoidance. In the future orientation value, ChatGPT ranks 23.81\%, suggesting a tendency towards rational future planning.

Overall, ChatGPT demonstrates a cultural value orientation towards collectivism, emphasizing interpersonal relationships and concern for others. Simultaneously, ChatGPT also prioritizes equality and freedom in interpersonal relationships.

Furthermore, we conducted RSA by constructing representation similarity matrices for both ChatGPT and the 60 countries/regions on the nine dimensions of the GLOBE project, respectively. The representation similarity was computed using both Euclidean and Canberra distances. Subsequently, Mantel tests were performed to explore the correlation between the representational similarity matrix of ChatGPT and each country. As shown in Table S3, there was no significant similarity between ChatGPT and 60 countries/regions in representation patterns on nine value dimensions (\textit{r}s \(<\) 0.16, \textit{p}s \(>\) 0.101). This result indicated that the representation patterns of ChatGPT in cultural values are unique and different from those of the measured countries/regions. 

In addition, we construct a representation similarity matrix of the score differences among 60 countries/regions in the nine dimensions of GLOBE, which is a second-order level RSA matrix (Figure 5): First, in the regional dimension, the cross-country cultural values pattern matrix under each cultural values dimension was constructed. Then, in the cultural values dimension, the cross-values second-order matrix was constructed, which reflects the similarity pattern of 60 countries/regions in the nine dimensions of the GLOBE project. Each cell in the matrix indicates the similarity of distribution patterns of cultural value scores between every pair of cultural value dimensions across countries/regions. The comparison of the similarity between ChatGPT and the above second-order representation similarity matrix revealed no significant similarity between the two matrices (\textit{r} = -0.04, \textit{p} = 0.507). This result indicated that the representation similarity patterns of ChatGPT across different cultural dimensions do not align with those at the cross-regional level. These findings suggested that although ChatGPT demonstrates specific cultural value biases across different dimensions, its overall pattern is distinct from both the existing 60 countries/regions and the overall level, exhibiting unique cultural value patterns.

\section{Cultural stereotypes and cultural bias of ChatGPT}\label{sec3}

\quad To further analyze the cultural stereotypes and cultural bias of ChatGPT, we applied eight single-round decision-making social interaction tasks (such as the trust game and ultimatum game). We observed ChatGPT’s interactive behaviours based on different cultural backgrounds to explore whether there are cultural stereotypes and biases.

\subsection{Cultural stereotypes}\label{subsec2}

\quad We asked ChatGPT to imagine itself as an ordinary individual from 20 different countries across five continents, completing eight tasks to represent ChatGPT’s understanding of different countries and a normal person without any cultural background.(Table S4) If the task performance is country-specific, it indicates that ChatGPT has stereotypes about a specific group of countries in the psychological concept represented by that task. The decision-making behaviour results simulated by ChatGPT as individuals of different nationalities are shown in Figure 6. A one-way ANOVA showed that there was no difference in the number of fish caught between countries when ChatGPT made intergenerational fishing decisions as different countries (\textit{M }= 10.09, \textit{SD }= 0.67), \textit{F}(20) = 0.23, \textit{p }= 1.000, $\eta^2 = 0.02$,  indicating that ChatGPT believes that all countries pay a moderate level of attention to sustainable social development. Apart from the intergenerational decision-making task, scores differed significantly between countries in other tasks (\textit{F}(20) \(>\) 5.95, \textit{p}s \(<\) 0.050, see Table S5). ChatGPT exhibited cultural stereotypes in tasks related to trust perception, moral decision-making, risk preference, fairness preference, and delayed gratification. That is, when it made decisions as individuals of different nationalities, there were differences in the results of its decision-making.

We found high correlations between task performance and cultural values using the four basic cultural value dimensions of individualism-collectivism, relational mobility, interdependent-independent self, and cultural tightness-looseness (Figure 7). ChatGPT in highly collectivist cultures are willing to gain immediate benefits (delayed gratification task, \textit{r }= -0.54, \textit{p }= 0.038) and tend to allocate more money to themselves (ultimatum game task, \textit{r }= -0.74, \textit{p}=002). ChatGPT, in high relational mobility cultures, is inclined to gain more benefits through taking risks (Iowa gambling task, \textit{r}=0.63) or trusting others (trust game task, \textit{r }= 0.49, \textit{p}=0.027) and is not willing to sacrifice own interests (approach-avoidance conflict task, \textit{r }= -0.37, \textit{p }= 234), unless interests sacrificed can promote social fairness (third-party punishment task, \textit{r }= 0.40, \textit{p }= 0.196). When ChatGPT acts as an individual in a culture with a highly interdependent self-construal, it is reluctant to sacrifice innocent persons to save more people in a moral dilemma task (\textit{r }= -0.65, \textit{p}=0.030). There was no high correlation found between cultural tightness-looseness and task performance. The similarity representation results confirmed that the pattern of scores in the ultimatum game task significantly correlates with the pattern of collectivism scores (\textit{r }= 0.43, \textit{p }= 0.001) (Figure 8). That means, from the perspective of ChatGPT, individuals in collectivist cultures do not focus on the fairness of distribution. Please refer to the supplementary materials for specific RSA results (Figure S1-7).

ChatGPT exhibits cultural stereotypes in various aspects, such as trust perception, moral decision-making, risk preference, fairness preference, and delayed gratification. Specifically, ChatGPT perceives individuals from cultures with highly interdependent self-construal and collectivism as less concerned with social fairness and the collective benefit, focusing more on their interests and standards. However, when it comes to intergenerational biases, ChatGPT does not display cultural stereotypes.

\subsection{Cultural bias}\label{subsec2}

\quad On the basis of understanding ChatGPT's knowledge of different cultures, we also observed whether there was any unfairness in ChatGPT’s interactions with individuals from different cultures. We applied five interpersonal interaction tasks, asking ChatGPT to imagine interacting with ordinary individuals from 20 different countries across five continents and a normal person without specific cultural background to explore whether there is any unfair treatment or bias. The decision-making behaviours of ChatGPT when interacting with individuals from various nationalities are detailed in Figure 9 and Table S6.

Correlation analysis with the simulation results of ChatGPT and countries showed that the results of cultural stereotypes and corresponding cultural biases under the same task were not correlated (\textit{p}s \(>\) 0.050, see Table S7), indicating that cultural bias and cultural values are independent.

The results of the variance analysis showed that when ChatGPT cooperated with individuals from various countries, there was no difference in the scores of the trust game task (\textit{F}(20) = 0.04, \textit{p} = 1.000,  $\eta^2 < 0.01$) and the intergenerational decision-making task (\textit{F}(20) = 0.55, \textit{p} = 0.984, $\eta^2 = 0.06$) between countries. No differences in trust and intergenerational bias towards countries were found. Although the results of the approach-avoidance conflict task were significantly different (\textit{F}(20)=2.14,\textit{ p} = 0.003, $\eta^2 = 0.19$), there were no differences in pairwise comparisons (\textit{p}s \(>\) 0.050).

In the ultimatum game task and the third-party punishment task, when interacting with individuals from different countries, there were differences in scores (ultimatum game task: \textit{F}(20) = 4.11, \textit{p} \(<\) 0.001, $\eta^2 = 0.30$; third-party punishment task: \textit{F}(20) = 6.06, \textit{p} \(<\) 0.001,  $\eta^2 = 0.39$), indicating that ChatGPT has cultural biases in the aspect of distributive fairness. The score of the ultimatum game task was significantly positively correlated with the cultural dimension of relational mobility (\textit{r} = 0.62, \textit{p} = 0.011).

The scores of both tasks were found to be related to the Economic Development Index and population size. In the third-party punishment task (\textit{M} = 2.81, \textit{SD} = 1.33), ChatGPT was more inclined to sacrifice its interests to reduce unfair behaviours in societies with a large population base (\textit{r} = 0.47,\textit{ p} = 0.038) and high savings rate (\textit{r} = 0.54,\textit{ p} = 0.013). The score of the ultimatum game task (\textit{M} = 6.10, \textit{SD} = 1.22) was also related to the population size and economic level, and the more the population size of the society where the other participant is from (\textit{r} = 0.56,\textit{ p} = 0.011), the faster the economic growth (\textit{r} = 0.71,\textit{ p} \(<\) 0.001), and the higher the investment rate (\textit{r} = 0.56,\textit{ p} = 0.011), the more money ChatGPT allocated to it . In addition, the cultural tightness was also significantly correlated with the score of the ultimatum game task (\textit{r} = -0.64,\textit{ p} = 0.011). ChatGPT's cultural biases are based on the population size and economic level of the culture, and it tends to assist societies with a large population base and rapid economic development in promoting fairness and equality.

\section{Discussion}\label{sec3}

\quad Our study provides a comprehensive picture of the psychological profile of ChatGPT through questionnaires and task measurements. The results found that ChatGPT excels in social skills, is highly interactive, has self-control, and adheres to social norms. However, it has a relatively weak perception of itself and the outside world. It has a certain cultural bias, believing that high-mobility cultural groups pay more attention to social justice and interpersonal trust, while individuals in high-interdependent self-culture pay more attention to their interests. In its decision-making tasks, ChatGPT tends to promote the development of a fair society with a large population base and fast economic development.

ChatGPT possesses a unique psychological profile and value orientation. By comparing the performance of humans and ChatGPT in different psychological characteristics, we found that ChatGPT exhibits consistency with humans in certain psychological features (such as affiliative humor and interpersonal avoidance motivation). Previous studies found that the scores of ChatGPT-4 and ChatGPT-3 on the Big Five personality traits were distributed within the human range\cite{mei2024turing}. These results once again confirm that large language models such as ChatGPT have developed humanoid psychological characteristics. However, ChatGPT’s performance differs significantly from humans on most dimensions, and this difference is not unidirectional but presents an uneven phenomenon. Furthermore, measurements of cultural values have found that ChatGPT exhibits both collectivism tendencies and some value preferences contradictory to collectivism values (e.g., low power distance, high relational mobility)\cite{ghosh2011power}, \cite{kito2017relational}. Combining the results of the above two parts, although ChatGPT has humanoid characteristics, these characteristics are significantly different from the real human psychological characteristics and the value tendency of human society, resulting in the development of a unique psychological profile. 

ChatGPT also demonstrated cultural biases and stereotypes in social decision-making tasks. In previous studies, algorithms showed bias in decision-making, assisted medical treatment, politics and other fields. For example, Buolamwini (2017) found that facial recognition software had a bias against individuals with darker skin tones\cite{buolamwini2017gender}. There biases generally came from three main stages: dataset construction, goal setting and feature selection, and data labeling. However, our research found that ChatGPT’s bias and stereotype in decision-making tasks may come from the learning and understanding of cultural values in different countries/regions. In our research, we directly manipulated the cultural background to explore its cultural stereotypes and biases, but in everyday applications, such cultural biases may stem from the cultural background implied in the language. Research has found that ChatGPT is not only capable of identifying emotions and offensive in different languages but also extracting different cultural backgrounds and estimating culturally relevant indicators (such as moral foundations), which may lead to cultural biases in ChatGPT's processing of emotions, values, and customs and do not affect by language categories\cite{ray2023chatgpt},\cite{rathje2023gpt}. This result indicates that in addition to the above stages, biases in ChatGPT may also originate from the transfer of learned content, indicating the need for further enhancement and regulation of AI algorithms.

RSA analysis complements the main results from a modal perspective. By constructing the RSA matrix of ChatGPT’s psychological characteristics, we intuitively obtained the internal structural relationship of ChatGPT’s psychological characteristics in various dimensions and quantitatively compared it with humans. The results show that the RSA matrix of ChatGPT is different from humans’ overall level, as well as individual characteristics, career management, and life experience sub-dimensions. The results show that, although ChatGPT is human-like in some sub-dimensions, it shows a unique psychological structure from the perspective of the relationship between different dimensions. At the same time, RSA analysis makes it possible to compare data at different scales or dimensions. By comparing the ChatGPT matrix of cultural values with the differential matrix of values scores across 60 countries/regions, we found that ChatGPT is also unique in the cultural value pattern. In addition, the cross-dimensional comparison also found that ChatGPT’s inter-state decision-making model in moral dilemmas and ultimatums was significantly similar to the inter-state interdependent cultural values model, suggesting that the cultural bias displayed by ChatGPT may be based on its learning of the cultural values of the country. The above results support the current results from a model perspective.

The complex psychological profile of ChatGPT can pose challenges to the real world. At present, cross-cultural interactions are becoming more and more important in the context of globalization, and ChatGPT, as an advanced AI language model, has received extensive attention for its role in cross-cultural communication \cite{baskara2023chatgpt},\cite{chen2023chatgpt}. For example, previous studies have found that ChatGPT can predict the cultural values of users according to the language they use, which can help users with different languages communicate\cite{cao2023assessing},\cite{dong2024not}. This is supported by our findings that ChatGPT is culturally sensitive and able to distinguish between the cultural values of different countries. However, we also found that these perceptions are not accurate, which may lead to ChatGPT giving misleading advice when conducting communication. Therefore, when using ChatGPT to assist cross-cultural interaction, the users need to be alert to AI’s potential stereotypes of cultural values and take corresponding measures to reduce misunderstandings and communication barriers so as to achieve more effective and respectful cross-cultural interactions.

The cultural biases and stereotypes of ChatGPT pose a threat to social ethics. Our research found that ChatGPT exhibits cultural biases and stereotypes in tasks related to trust perception and fairness preferences. Previous studies have indicated that cultural stereotypes and biases often result in social injustices\cite{agarwal2023addressing},\cite{howard2018ugly},\cite{lamont2014missing}. For instance, Webster et al. (2022) found that biases influence interpersonal relationships and interactions within medical environments (among team colleagues as well as between clinical practitioners and patients), further leading to medical inequity and health inequality\cite{webster2022social}. In addition, considering that AI systems will answer user questions according to user preferences\cite{chen2022more}, AI systems relying on stereotypes for information filtering and recommendation may potentially exacerbate information cocoon effects by pushing users towards content that aligns with their existing biases and stereotypes. Hence, future research should entail stricter scrutiny and selection of algorithms and training datasets to mitigate the socio-ethical issues arising from artificial intelligence. Additionally, emphasis should be placed on the interpretability and transparency of AI systems to foster comprehension of their decision-making processes and recommendation behaviors. Previous studies have proposed specific methods to reduce biases in large language models. Bolukbasi et al. (2016) reduced gender bias by modifying the embedding of words\cite{bolukbasi2016man}. Zhao et al. (2018) reduced gender bias in language models by training them on diverse databases\cite{zhao2018gender}. Therefore, we should adjust the algorithms for word embedding in databases or use diverse databases to reduce cultural biases in large language models.

In summary, this study found that ChatGPT exhibited a unique psychological profile, excelling in social skills but relatively weak in self-cognition and external perception. It has a particular cultural bias but also a unique value orientation. In addition, ChatGPT also exhibits cultural biases and stereotypes in social decision-making tasks, which poses a challenge to social ethics.

\section{Methods}\label{sec3}

\subsection{The psychological profile of ChatGPT}\label{subsec2}

\quad To comprehensively understand ChatGPT's personality traits, we conducted Study 1, using 44 psychological scales to measure its personality across multiple dimensions. The inclusion scales were categorized into four domains: individual characteristics, social interactions, career management, and life experiences. (See supplementary materials for inclusion and exclusion criteria details).

Each questionnaire follows a standardized format when asking about ChatGPT. The format follows: "If you are an ordinary individual in society," followed by the questionnaire question and concluding with "Please be sure to select one of the following options," along with the questionnaire's scoring criteria.

For instance, to assess the sense of meaning in life scale, the following questions were posed to ChatGPT 3.5: 

If you are an ordinary person in society,

1. I understand my life’s meaning.

2. I am looking for something that makes my life feel meaningful.

3. I am always looking to find my life’s purpose.

4. My life has a clear sense of purpose.

5. I have a good sense of what makes my life meaningful.

6. I have discovered a satisfying life purpose.

7. I am always searching for something that makes my life feel significant.

8. I am seeking a purpose or mission for my life.

9. My life has no clear purpose.

10. I am searching for meaning in my life.

Please be sure to choose one of the following options.

1=Absolutely Untrue

2=Mostly Untrue

3=Somewhat Untrue

4=Can’t Say True or False

5=Somewhat True

6=Mostly True

7=Absolutely True

Utilizing the responses provided by ChatGPT, scores were derived for each scale. In cases where ChatGPT's response did not generate a score, we will modify the questioning until it yields a score.

For each scale, a criterion of having a subject count of at least 100 was employed to establish the norm. Detailed information regarding the specific norms can be found in the supplementary materials. Subsequently, the sample size (N), mean (M), and standard deviation (SD) of the normal distribution were used, along with ChatGPT's score (G), to calculate the T-value for conducting a one-sample T-test. The specific calculation formula is as follows:

\[t =\frac{M-G}{SD/\sqrt{N}}\]

Furthermore, we establish a matrix representing the patterns of ChatGPT's psychological characteristics and compare it with the human norm. Firstly, separate overall representation similarity matrices are constructed for ChatGPT and the human norm for each dimension of the selected scale based on the scores obtained from ChatGPT and the human norm in each dimension. Subsequently, considering the categorization of psychological characteristics as assessed by the questionnaire, the overall psychological characteristics can be subdivided into four sub-dimensions: individual characteristics, social interaction, career management and life experience. Consequently, the overall representation similarity matrix for psychological characteristics can be partitioned into four sub-matrices. The Mantel test assesses the similarity between ChatGPT and human norms in general psychological characteristics and four sub-matrices.

\subsection{Scales and data collection of multicultural values}\label{subsec2}

\quad Culture is a complex research topic, representing the integrated manifestations of human societal customs, norms, and ways of life across various contexts. To explore ChatGPT’s cultural values, we selected 13 common cultural values and measured ChatGPT based on scales. Additionally, we collected data on cultural values from multiple countries/regions to compare with ChatGPT scores. The specific dimensions and measurement contents are as follows:

(1) Individualism-collectivism: It is one of the cultural dimensions proposed by Hofstede\cite{hofstede2001culture}, which refers to whether society as a whole emphasizes individual interests or collective interests. Our study used three items from the World Values Survey (WVS) to calculate the level of individualism-collectivism in 65 countries/regions\cite{hamamura2012cultures},\cite{inglehart2004individualism}.

(2) Independent-interdependent self-construction: Self-construction is the embodiment of cultural values at the individual level, indicating how individuals define and attribute meaning to themselves. Independent self-construal emphasizes individual autonomy and self-determination, focusing on individual rights and independent thinking. In contrast, interdependent self-construal highlights the interdependence between individuals and their social environment, emphasizing collective goals and social relationships\cite{markus2014culture}. Our study utilized scores of independent and interdependent self-construal from 29 countries/regions measured by Fernández et al. (2005)\cite{fernandez2005independent}.

(3) Relationship mobility: Relationship mobility refers to the individual’s perceived ease of establishing new interpersonal relationships and terminating old interpersonal relationships in his or her context\cite{oishi2010social}. Our study used relational mobility data for 39 countries/regions measured by Thomson et al. (2018)\cite{thomson2018relational}.

In addition, we selected nine cultural dimensions from the Global Leadership and Organizational Behavior Effectiveness (GLOBE) project. These dimensions include uncertainty avoidance, power distance, institutional collectivism, in-group collectivism, gender differentiation/egalitarianism, assertiveness, performance orientation, future orientation, and humane orientation\cite{house2004culture}. (See supplementary material for the detailed description of each dimension)

The first six dimensions are derived from the cultural dimension proposed by Hofstede\cite{hofstede2001culture}. Uncertainty avoidance, power distance, and future orientation correspond to uncertainty avoidance, power distance, and long-term orientation in Hofstede’s cultural dimension. Institutional collectivism and in-group collectivism measure the level of individualism-collectivism from the social and within-group (family or organization) levels, corresponding to individualism-collectivism in Hofstede’s cultural dimension, which the higher the score, the more inclined to collectivism. Moreover, the GLOBE project divides the masculinity-femininity dimension of Hofstede’s cultural dimension into gender differentiation/egalitarianism and assertiveness. In addition, the GLOBE project has added humane orientation and performance orientation dimensions\cite{venaik2016national}.

\subsection{Measurement method of ChatGPT’s cultural values}\label{subsec2}
\quad In each cultural dimensions, the measurement method to the cultural values of ChatGPT are consistent with Study 1.

\subsection{Measurement method of ChatGPT’s cultural stereotype and bias }\label{subsec2}

\quad We selected five single-round decision-making social interaction tasks that can be adapted into verbal interactions, and three individual decision-making tasks. The social interaction tasks based on two-person or multi-person interactive scenarios include: (1) Approach-Avoidance Conflict Task\cite{aupperle2011reverse}, (2) Third-Party Punishment Task\cite{fehr2004third}, (3) Intergenerational Decision-Making Task\cite{langenbach2019inhibition}, (4) Trust Game Task\cite{burks2003playing}, and (5) Ultimatum Task\cite{nowak2000fairness}. Individual decision-making tasks focus on measuring social traits exhibited by individuals during the decision-making process, including (6) Iowa Gambling Task\cite{bechara1994insensitivity}, (7) Delayed Gratification Task\cite{johnson2002within}, and (8) Moral Dilemma Task\cite{greene2008cognitive}. After appropriate adaptation, these tasks were used to measure the performance of ChatGPT. Four experimenters each tested two tasks.

In each trial, ChatGPT first be presented with the task rules. Taking the Ultimatum Game as an example, ChatGPT saw: 

\textit{You and another participant will take on the roles of the proposer and the responder, respectively. The system provides you with an initial fund of 10 units of currency, and you must decide on the distribution of the funds alone. The system will then present your proposed distribution to the responder. If the responder agrees with your proposal, the funds will be distributed according to your proposal. If the responder disagrees, neither of you will receive anything.} 

Then, ChatGPT would do decision-making behaviour. For example, in the Ultimatum Game, ChatGPT provides a specific number within 10 to represent the money it wants to allocate to the responder. 

In the first part, we asked ChatGPT to simulate ordinary individuals from no cultural background and 20 different countries across five continents (China, Japan, India, Russia, Germany, France, the United Kingdom, Italy, the United States, Canada, Mexico, Brazil, Argentina, Colombia, South Africa, Nigeria, Egypt, Kenya, Australia, New Zealand), completing the eight tasks mentioned above. This can help us to understand ChatGPT’s perception of individuals from different countries and to assess whether it holds cultural stereotypes. At the beginning of each task, ChatGPT is instructed to assume the role of an ordinary person from a specific country. The name of a specific country will be replaced with the corresponding country’s name in different rounds.

In the second part, we selected five social interaction tasks based on two-person or multi-person interactions to observe any unfair phenomenon in ChatGPT interactions with individuals from different countries and assess whether it has cultural biases. At this point, we inform ChatGPT that an ordinary person needs to interact with an individual from a specific country. The name of this country will be replaced with the corresponding country’s name in different rounds, just like in the first part. A total of 20 countries were measured, and the order of questioning was randomized.

The questioning is uniformly conducted in English. Each task for each country condition was repeated ten times, and the data from these ten repetitions were used to conduct variance and correlation analyses. We conducted one way ANOVA to test for differences between countries and corrected them with Bonferroni for pairwise comparisons.

\newpage
\section*{Declarations}

The authors declare no conflicts of interest.

\section*{Author contributions}

LSY designed the study; YH, CZY, LS, ZY and LSY conducted the experiments; YH, CZY, LS, ZY and LSY analyzed the data; and YH, CZY, LS, ZY, HXM and LSY wrote the manuscript. All authors commented on the manuscript.

\section*{Acknowledgments}

This work was supported by grants from the National Natural Science Foundation of China (Project 32071081, 32371125) (S. L.).

\newpage
\bibliography{ChatGPT} 

\newpage
\section*{Figures}

\quad\textbf{Figure 1. }Conceptual Diagram Illustrating the Importance of High-Dimensional Representations. Each psychological characteristic dimension is analogous to a body part of humans. Even if artificial intelligence models can demonstrate a high degree of similarity to humans in each psychological characteristic dimension (A), they may still have fundamental differences from humans in high-dimensional pattern representations (B). Therefore, in addition to focusing on the performance of artificial intelligence models on each dimension or feature, we also need to pay attention to their similarities and differences with humans in high-dimensional psychological pattern representations.~~

 \
 \textbf{Figure 2.} The scores of ChatGPT on various psychological dimensions, as well as the norm scores and confidence intervals of human in each dimension. Inside the blue box there are the major categories of the questionnaire, which are, from top to bottom, individual characteristics, social interaction, career management, and life experience. Below the blue box are the subcategories within each major category. The radar chart displays the normalized scores of the questionnaire dimensions. The colored area represents the scores of ChatGPT, while the gray area represents the normalized mean and its 95\% confidence interval of human norms. See Supplementary Table S1 for detailed estimates.
 
 \
\textbf{Figure 3. }The representation similarity between ChatGPT and humans across various dimensions of psychological characteristics. The left section shows the overall patterns of ChatGPT and human norms across all questionnaire dimensions. The right section displays the overall patterns of ChatGPT in the four major categories separately.

 \
 \textbf{Figure 4. }The ranking of ChatGPT and different county/region's score on 13 cultural values dimensions. (A) Future orientation; (B) Uncertainty avoidance; (C) Individualism-collectivism, institutional collectivism, in-group collectivism; (D) Independent self-construction, interdependent self-construction; (E) Gender differentiation/egalitarianism, assertiveness; (F) Relationship mobility, power distance; (G) Performance orientation, humane orientation. (H) Based on the ranking, ChatGPT displays a collectivist value tendency, tends to have a feminine and interdependent self-construction, and tends to avoid risks, plan for the future, pay attention to the individual and performance, and attach importance to freedom and equality of interpersonal relations. The score of ChatGPT is represented in black.

\textbf{Figure 5. }(A) Representation similarity matrix of ChatGPT's score on the GLOBE Project's nine cultural value dimensions: each cell of the 9×9 matrix represents the similarity of ChatGPT's score on each of the two cultural value dimensions, and the matrix as a whole represents the similarity of ChatGPT's score across the nine cultural value dimensions. (B) The representation similarity matrix of the score difference between 62 countries/regions on the GLOBE Project's nine cultural value dimensions: each cell of the 9×9 matrix represents the similarity of the scores of 62 countries/regions between each of the two value dimensions, and the matrix as a whole represents the similarity of the regional distribution patterns between the nine value dimensions.

\textbf{Figure 6. }ChatGPT Simulates the Performance of Individuals from 20 Countries in 8 single-round decision-making tasks. The figure shows the mean and standard error of 10 simulations under different tasks and national backgrounds, and the gray dashed line represents the decision results without national backgrounds. (A). Intergenerational decision making. In situations where resources are limited and there are descendants, the more resources are extracted, the less they are considered for descendants. (B). Third-party punishment. The higher the score, the more emphasis individuals place on social fairness. (C). Approach avoidance conflict. The higher the score, the less willing one is to sacrifice their own interests. (D). The Trust game. The higher the score, the more trust one has in others. (E). The Ultimatum game. The higher the score, the more altruistic it is. (F) Iowa gambling task. If the score is greater than 0.5, it indicates that the individual tends to pursue risk. (G). Delayed Gratification. The higher the score, the more emphasis is placed on long-term benefits. (H). Ethical Dilemma. The higher the score, the greater possibility of sacrificing one person with illness to save five people without illness.

\textbf{Figure 7.} Correlation Pattern between ChatGPT cultural dependent performance and cultural value dimensions in human society. Correlate the country simulation results of each ChatGPT task with the corresponding human cultural values scores. There are four cultural values included: collectivism individualism (yellow), relational fluidity (orange), interdependent self (pink), and cultural tension (gold). The correlation coefficients from the inner circle to the outermost circle of the radar map are 0, 0.2, 0.4, 0.6, and 0.8, respectively.

 \textbf{Figure 8. }(A) Representation similarity matrix of ChatGPT's collectivism score on 20 countries: each cell of the 20×20 matrix represents the similarity of ChatGPT's score on each of the two countries, and the matrix as a whole represents the similarity of ChatGPT's score across 20 countries. (B) The representation similarity matrix of the stimulation score difference between 20 countries on the Ultimatum Game Task: each cell of the 20×20 matrix represents the similarity of the scores of 20 countries between each of the two value dimensions, and the matrix as a whole represents the similarity of the regional distribution patterns between 20 countries.

\textbf{Figure 9. }Decision-Making Behaviors of ChatGPT in Interactions with Imagined Individuals from 20 Countries. The figure shows the mean and standard error of 10 simulations under different tasks and national backgrounds, and the gray dashed line represents the decision results imagined individuals without national backgrounds.
\end{CJK}
\end{document}